\def\year{2018}\relax
\documentclass[letterpaper]{article} 
\usepackage{aaai18}  
\usepackage{times}  
\usepackage{helvet}  
\usepackage{courier}  
\usepackage{url}  
\usepackage{graphicx}  
\usepackage{mathtools}
\usepackage{algorithm}
\usepackage[noend]{algorithmic}
\usepackage{booktabs}
\usepackage{amsmath}
\usepackage{amssymb}
\usepackage{amsthm}

\newcommand{\pop}[1]{\Delta_{#1}}
\newcommand{\lv}[1]{#1}
\newcommand{\lvs}[1]{\mathbf{#1}}
\newcommand{\ind}[1]{#1}
\newcommand{\inds}[1]{\mathbf{#1}}
\newcommand{\const}[1]{#1}
\newcommand{\pred}[1]{\ensuremath{\mathsf{#1}}}
\newcommand{\true}{\ensuremath{\mathsf{True}}}
\newcommand{\false}{\ensuremath{\mathsf{False}}}

\newcommand{\atoms}[1]{\ensuremath{\mathcal{#1}}}
\newcommand{\grounds}[1]{\ensuremath{\mathcal{G(#1)}}}

\newcommand{\fig}[1]{Fig~\ref{#1}}
\newcommand{\WF}[2]{\ensuremath{\left\langle #1, #2 \right\rangle}}
\newcommand{\counter}[2]{\ensuremath{\eta(#1,#2)}}

\newcommand{\Din}{D_{\hat{O}}}
\newcommand{\Dout}{D_{\hat{I}}}

\usepackage{amsmath}
\usepackage{amssymb}
\usepackage{amsthm}

\newtheorem{defn}{Definition}

\newtheorem{exampl}{Example}
\newenvironment{example}{\begin{exampl}\em}{\end{exampl}}
\newtheorem{analog}{Analogy}

\newtheorem{motiv}{Motivation}
\newenvironment{motivation}{\begin{motiv}\em}{\end{motiv}}

\begin{document}
\title{RelNN: A Deep Neural Model for Relational Learning}
\author{Seyed Mehran Kazemi \and David Poole\\
University of British Columbia\\
Vancouver, Canada\\
\{\texttt{smkazemi, poole}\}@cs.ubc.ca\\
}
\maketitle
\begin{abstract}
Statistical relational AI (StarAI) aims at reasoning and learning in noisy domains described in terms of objects and relationships by combining probability with first-order logic. 
With huge advances in deep learning in the current years, combining deep networks with first-order logic has been the focus of several recent studies. Many of the existing attempts, however, only focus on relations and ignore object properties. The attempts that do consider object properties are limited in terms of modelling power or scalability.
In this paper, we develop relational neural networks (RelNNs) by adding hidden layers to relational logistic regression (the relational counterpart of logistic regression). We learn latent properties for objects both directly and through general rules. Back-propagation is used for training these models. A modular, layer-wise architecture facilitates utilizing the techniques developed within deep learning community to our architecture.
Initial experiments on eight tasks over three real-world datasets show that RelNNs are promising models for relational learning.\footnote{Code: \url{https://github.com/Mehran-k/RelNN}}
\end{abstract}

\section{Introduction}
Multi-relational data cannot be directly fed into traditional machine learning approaches such as logistic regression, SVMs, random forests, etc. 
Statistical relational AI (StarAI) \cite{StarAI-Book} aims at developing models that work directly with relational data by capturing the interdependence among properties of the objects and the relationships they are involved in.

In relational data, there are usually several classes of objects, each class has certain properties defined for its members, and there are (various) relationships among the objects. 
Many works on learning from relational data focus only on predicting new relationships among objects given known relationships, and ignore predicting object properties.
For the works that do consider predicting object properties, a recent comparative study shows that none of them perform well and that this problem is still only poorly understood \cite{kazemi2017comparing}. In this paper, we focus (primarily) on predicting a property of the objects in one class based on the rest of the data (e.g., predicting the gender of people given the movies they like). This problem is challenging when the property of each object in the class depends on a varying number of other objects' properties and relationships. In the StarAI community, this problem is known as \emph{aggregation}.

With numerous advances in learning deep networks for many different applications, using deep neural approaches for relational data has been the focus of several recent studies. In this paper, we develop a framework for learning first-order deep neural models to learn from and reason with relational data. 
Our model is developed through deepening relational logistic regression (RLR) models \cite{Kazemi:2014}, the directed analogue of Markov logic \cite{Richardson:2006aa}, by enabling them to learn latent object properties both directly and through general rules, and connecting multiple RLR layers as a graph to produce relational neural networks (RelNNs). Similar to \cite{niepert2016learning,pham2017column}, we identify the relationship between our model and convolutional neural networks (ConvNets). Identifying this relationships allows RelNNs to be understood and implemented using well-known ConvNet primitives.
Each training iteration of RelNNs is order of the amount of data times the size of the RelNN, making RelNNs highly scalable.

We evaluate RelNNs on three real-world datasets and compare them to well-know relational learning algorithms. 
We show how RelNNs address a relational learning issue raised in \cite{Poole:2014} who showed that as the population size increases, the probabilities of many variables go to $0$ or $1$, making the model over-confident. 
Obtained results indicate that RelNNs are promising models for relational learning.

\section{Relational Logistic Regression and Markov Logic Networks}
StarAI models aim at modelling the probabilities about relations among objects. Before knowing anything about the objects, these models treat them identically and apply tied parameters to them. In order to describe these models, first we need to introduce some definitions and terminologies.

A {\bf population} is a finite set of {\bf objects}. 
\textbf{Logical variables} (logvars) start with lower-case letters, and \textbf{constants} denoting objects start with upper-case letters.  
Associated with a logvar $\lv{x}$ is a population $\pop{x}$ with $|\pop{x}|$ representing the cardinality of the population. 
A lower-case letter in bold refers to a tuple of logvars. 
An upper-case letter in bold refers to a tuple of constants.
An \textbf{atom} is of the form $\pred{S}(t_1, \dots, t_k)$ where $\pred{S}$ is a predicate symbol and each $t_i$ is a logvar or a constant.
We write a \textbf{substitution} as $\theta=\{\langle \lv{x_1}, \dots, \lv{x_k} \rangle \slash \langle t_1, \dots, t_k \rangle\}$ where each $\lv{x_i}$ is a different logvar and each $t_i$ is a logvar or a constant in $\pop{x_i}$.
A \textbf{grounding} of an atom with logvars $\{\lv{x_1}, \dots, \lv{x_k}\}$ is a substitution $\theta=\{\langle \lv{x_1}, \dots, \lv{x_k} \rangle \slash \langle \const{X_1}, \dots, \const{X_k} \rangle\}$ mapping each of its logvars $\lv{x_i}$ to an object $\const{X_i} \in \pop{x_i}$. Given a set $\atoms{A}$ of atoms, we denote by $\grounds{A}$ the set of all possible groundings for the atoms in $\atoms{A}$.
A \textbf{literal} is an atom or its negation.
A \textbf{formula} $\varphi$ is a literal, a disjunction $\varphi_1\vee\varphi_2$ of formulas or a conjunction $\varphi_1\wedge\varphi_2$ of formulas.
A \textbf{conjunctive formula} has no disjunctions.
Applying a \textbf{substitution} $\theta=\{\langle \lv{x_1}, \dots, \lv{x_k} \rangle \slash \langle t_1, \dots, t_k \rangle\}$ on a formula $\varphi$ (written as $\varphi\theta$) replaces each $\lv{x_i}$ in $\varphi$ with $t_i$.
A {\bf weighted formula (WF)} is a tuple $\WF{\varphi}{w}$ where $\varphi$ is a formula and $w$ is a weight.

Let $\pred{Q}(\lvs{x})$ be an atom whose probability depends on a set $\atoms{A}$ (not containing $\pred{Q}$) of atoms (called the parents of \pred{Q}), $\psi$ be a set of WFs containing only atoms from $\atoms{A}$, $\hat{I}$ be a function from groundings in $\grounds{A}$ to truth values, $\inds{X}$ be an assignment of objects to $\lvs{x}$, and $\theta=\{\lvs{x} \slash \inds{X}\}$. 
{\bf Relational logistic regression (RLR)} \cite{Kazemi:2014} defines the probability of $\pred{Q}(\inds{X})$ given $\hat{I}$ as follows:
\begin{align}\label{RLR-EQ}
Prob(\pred{Q}(\inds{X})=True \mid \hat{I}) = sigmoid\big(sum \big)
\end{align}
where,
\begin{align}\label{RLR-SUM}
sum = \sum_{\WF{\varphi}{w} \in \psi}{w * \counter{\varphi\theta}{\hat{I}}}
\end{align}
where $\counter{\varphi\theta}{\hat{I}}$ is the number of instances of $\varphi\theta$ that are true w.r.t. $\hat{I}$. 
Note that $\counter{\true}{\hat{I}}=1$. Following \cite{Kazemi:2014}, w.l.o.g we assume the formulae of all WFs for RLR models are conjunctive.

Markov logic networks (MLNs) \cite{Richardson:2006aa} use WFs to define a probability distribution over ground atoms.
As shown in \cite{Poole:2014}, when all groundings in $\grounds{A}$ are observed, an RLR model is identical to an MLN with corresponding WFs. We use RLR/MLN when the two models are identical.

\begin{figure}
\begin{center}
\includegraphics[width=\columnwidth]{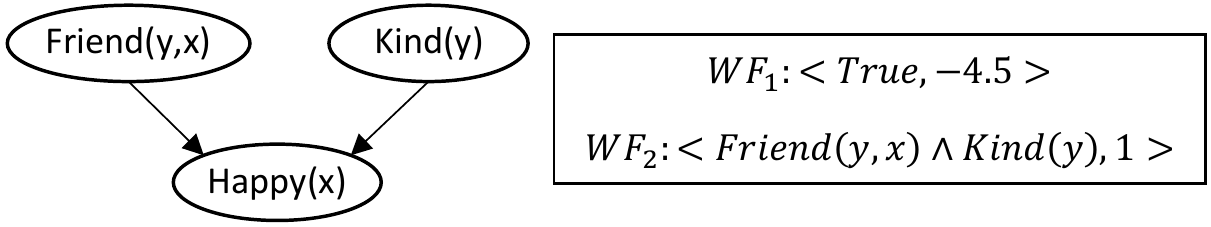}
\end{center}
\caption{An RLR model taken from \cite{Kazemi:2014}.}
\label{KindFriend}
\end{figure}

\begin{example}\label{happy-example}
Let \pred{Happy}(\lv{x}) be an atom which depends on $\atoms{A}=\{\pred{Friend}(\lv{x},\lv{y}), \pred{Kind}(\lv{y})\}$, and let $\hat{I}$ be a function from \grounds{A} to truth values. Let an RLR/MLN model define the conditional probability of \pred{Happy} using WFs in \fig{KindFriend}.
According to this model:
$Prob(\pred{Happy}(\ind{X})=True \mid \hat{I}) = sigmoid(-4.5 + 1*\counter{\pred{Friend}(\lv{y},\ind{X})\wedge \pred{Kind}(\lv{y})}{\hat{I}})$, 
where $\counter{\pred{Friend}(\lv{y},\ind{X})\wedge \pred{Kind}(\lv{y})}{\hat{I}}= \#\ind{Y} \in \pop{y}$ s.t. $\pred{Friend}(\ind{Y},\ind{X}) \wedge \pred{Kind}(\ind{Y})$ according to $\hat{I}$, corresponding to the number of friends of $\ind{X}$ that are kind. When this count is greater than or equal to 5, the probability of $\ind{X}$ being happy is closer to one than zero; otherwise, the probability is closer to zero than one. Therefore, the two WFs model ``someone is happy if they have at least 5 friends that are kind''. 
\end{example}

\textbf{Continuous atoms:} The RLR model defined above only works with Boolean or multi-valued parents. In order to allow for parents with continuous atoms, we use \citeauthor{fatemi2016learning} (\citeyear{fatemi2016learning})'s proposal. If \true\ and \false\ are associated with $1$ and $0$, $\wedge$ in WFs can be substituted with $*$. Then continuous atoms may be used in WFs. For example if for some $\ind{X} \in \pop{x}$ we have $\hat{I}(\pred{R}(\ind{X}))=1$, $\hat{I}(\pred{S}(\ind{X}))=0$, and $\hat{I}(\pred{T}(\lv{X}))=0.2$, then \WF{w}{\pred{R}(\ind{X}) *\pred{S}(\ind{X})} evaluates to $0$, \WF{w}{\pred{R}(\ind{X}) * \neg \pred{S}(\ind{X})} evaluates to $w$, and \WF{w}{\pred{R}(\ind{X}) * \neg \pred{S}(\ind{X}) * \pred{T}(\ind{X})} evaluates to $0.2*w$.

\begin{figure*}[t]
\begin{center}
\includegraphics[width=0.9\textwidth]{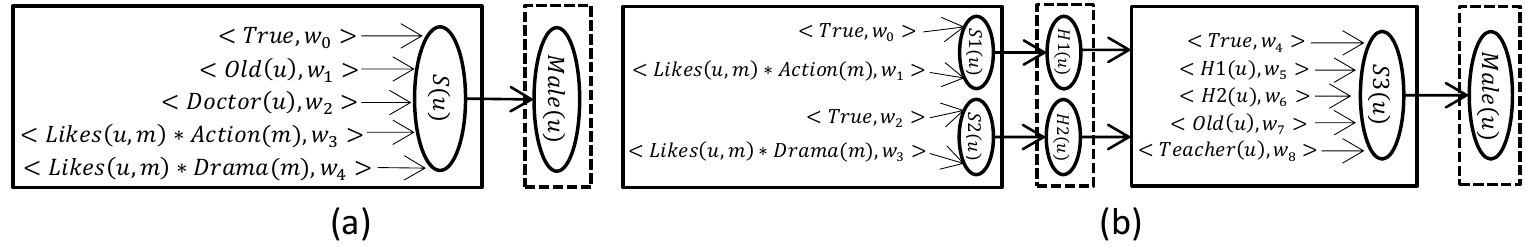}
\end{center}
\caption{An RLR and a RelNN models for predicting the gender of users in a movie rating system with a layer-wise architecture.}
\label{RLR-RelNN}
\end{figure*}

\section{Relational Neural Networks}
We encode RLR/MLN models with a layer-wise architecture by designing relational counterparts of the linear layer (LL), activation layer (AL), and error layer (EL) in neural networks. This enables building relational neural nets (RelNNs) by connecting several relational LLs and relational ALs as a graph. It also enables utilizing the \emph{back propagation} algorithm for training these models by adding a relational EL at the end of the graph, feeding the data forward, calculating the prediction errors, back propagating the derivatives with respect to errors, and updating the parameters.

Let $\atoms{A_{\mathnormal{in}}}$ and $\atoms{A_{\mathnormal{out}}}$ represent two distinct sets of atoms, and $E$ represent an error function. 
Let $\hat{I}$ be a function from $\pred{P}(\inds{X})\in\grounds{A_{\mathnormal{in}}}$ to values. Let $\hat{O}$ be a function from $\pred{Q}(\inds{X})\in\grounds{A_{\mathnormal{out}}}$ to values. Let $\Din$ be a function from $\pred{Q}(\inds{X})\in\grounds{A_{\mathnormal{out}}}$ to $\frac{\partial E}{\partial \pred{Q}(\inds{X})}$. Let $\Dout$ be a function from $\pred{P}(\inds{X})\in\grounds{A_{\mathnormal{in}}}$ to $\frac{\partial E}{\partial \pred{P}(\inds{X})}$.

A \emph{layer} is a generic box that given $\hat{I}$ as input, outputs $\hat{O}$ based on $\hat{I}$ and its internal structure, and given $\Din$, updates its internal weights and outputs $\Dout$ using the chain rule.

A \emph{relational linear unit} for an atom $\pred{Q}(\lvs{x})$ is identical to the linear part of RLR in Eq.~(\ref{RLR-SUM}).
A \emph{relational LL (RLL)} consists of $t$ $>$ $0$ \emph{relational linear units} with the $j$-th unit having atom $\pred{Q_j}(\lvs{x_j})$ and a set $\psi_j=\{\WF{w_{j1}}{\varphi_{j1}}, \WF{w_{j2}}{\varphi_{j2}}, \dots, \WF{w_{jm_j}}{\varphi_{jm_j}}\}$ of $m_j$ WFs. 
$\atoms{A_{\mathnormal{in}}}$ for an RLL contains all atoms in $\psi_j$s and $\atoms{A_{\mathnormal{out}}}$ contains all $\pred{Q_j}(\lvs{x_j})$s. $\hat{I}$ comes from input data or from the layers directly connected to the RLL.
For any $\pred{Q}(\inds{X})\in\grounds{A_{\mathnormal{out}}}$, $\hat{O}(\pred{Q}(\inds{X}))$ is calculated using the linear part of  Eq.~(\ref{RLR-EQ}) with respect to $\hat{I}$ and $\psi_j$. 
An RLL can be seen as general rules with tied parameters applied to every object. 

\begin{example}
Consider an RLL with $\atoms{A_{\mathnormal{in}}}=\{\pred{Friend(\lv{x},\lv{y})},\pred{Kind(\lv{y})}\}$, $\atoms{A_{\mathnormal{out}}}=\{\pred{Happy}(\lv{x})\}$, and with the WFs in Example~\ref{happy-example}.
Suppose there are $10$ objects: $\pop{x}=\{\const{X_1}, \const{X_2}, \dots, \const{X_{10}}\}$. Let $\hat{I}$ be a function from $\grounds{A_{\mathnormal{in}}}$ to values according to which $\const{X_1}, \const{X_2}, \dots, \const{X_{10}}$ have $3$, $0$, $\dots$, and $7$ friends that are kind respectively. 
$\hat{O}$ for this RLL is a function from groundings in $\grounds{A_{\mathnormal{out}}}$ to values as $\hat{O}(\pred{Happy}(\const{X_1})) \rightarrow -1.5, \hat{O}(\pred{Happy}(\const{X_2})) \rightarrow -4.5, \dots, \hat{O}(\pred{Happy}(\const{X_{10}}))\rightarrow 2.5$.
\end{example}

Consider the $j$-th relational linear unit.
For each assignment of objects $\inds{X_{ju}}$ to $\lvs{x_j}$, let $\theta_{ju}=\{\lvs{x_j}\slash\inds{X_{ju}}\}$.
The derivative with respect to each weight $w_{jk}$ can be calculated as $\sum_{\inds{X_{ju}}} \counter{\varphi_{jk}\theta_{ju}}{\hat{I}} * \Din(Q_j(\inds{X_{ju}}))$.
To show how $\Dout$ is calculated, for ease of exposition we assume no predicate appears twice in a formula (i.e. no self-joins). 
It is straight-forward to relax this assumption.
Let $\varphi\backslash \pred{P}$ represent formula $\varphi$ with any atom with predicate $\pred{P}$ (or its negation) removed from it.
For each assignment of objects $\inds{X_{iv}}$ to the logvars $\lvs{x_{i}}$ of \pred{P}, let $\theta_{iv}=\{\lvs{x_i}\slash\inds{X_{iv}}\}$.
Then $\Dout(\pred{P}(\inds{X_{iv}}))=\sum_{j}\sum_{\inds{X_{ju}}}\sum_{k=1,\pred{P}\in \varphi_{jk}}^{m_j} w_{jk} * \counter{((\varphi_{jk}\backslash \pred{P})\theta_{ju})\theta_{iv}}{\hat{I}} * \Din(\pred{Q_j}(\inds{X_{ju}}))$.

The \emph{relational AL (RAL)} and \emph{relational EL (REL)} are similar to their non-relational counterparts. RAL applies an activation function $A$ (e.g., $sigmoid$, $tanh$, or $ReLU$) to its inputs and outputs the activations. REL compares the inputs to target labels, finds the prediction errors based on an error function $E$, and outputs the prediction errors made by the model for each target object.

\begin{figure*}[t]
\begin{center}
\includegraphics[width=0.85\textwidth]{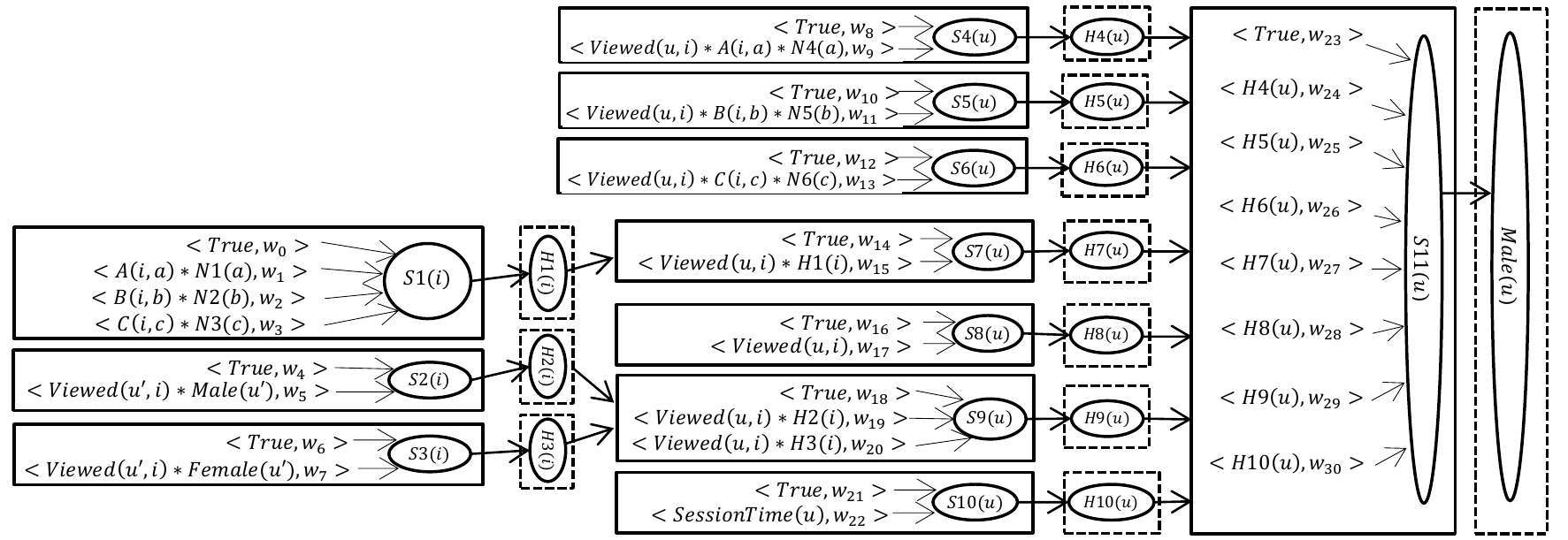}
\end{center}
\caption{RelNN structure for predicting the gender in PAKDD15 dataset.}
\label{KDD15-RelNN}
\end{figure*}

A \textbf{relational neural network (RelNN)} is a structure containing several RLL and RALs connected to each other as a graph. In our experiments, we consider the activation function to be the sigmoid function. During training, an REL is also added at the end of the sequence. 
\fig{RLR-RelNN} represents an example of an RLR and a simple RelNN model for predicting the gender of users based on their age, occupation, and the movies they like. We use boxes with solid lines for RLLs and boxes with dashed lines for RALs. For the first RLL, $\atoms{A_{\mathnormal{in}}}=\{\pred{Likes}(\lv{u},\lv{m}), \pred{Action}(\lv{m}), \pred{Drama}(\lv{m})$\} and all values in $\hat{I}$ come from the observations. For this RLL, $\atoms{A_{\mathnormal{out}}}= \{\pred{S1}(\lv{u}), \pred{S2}(\lv{u})\}$. For the second RLL, $\atoms{A_{\mathnormal{in}}}=\{\pred{Old}(\lv{u}),\pred{Teacher}(\lv{u}),\pred{H1}(\lv{u}),\pred{H2}(\lv{u})\}$ and $\hat{I}$ for the first two atoms comes from observations, and for the last two comes from the layers directly connected to the RLL.
\fig{KDD15-RelNN} shows a more complicated RelNN structure for gender predicting in PAKDD-15 competition.

\subsection{Motivations for hidden layers}

\begin{motivation} \label{motiv1}
Consider the model in \fig{RLR-RelNN}(a) and let $n_\const{U}$ represent the number of action movies that user $\const{U}$ likes. 
As \citeauthor{Poole:2014} (\citeyear{Poole:2014}) point out, if the probability of maleness depends on $n_\const{U}$, as $n_\const{U}$ increases the probability of maleness either goes to $0$ or $1$. 
This causes the model to become over-confident of its predictions when there exists many ratings and do a poor job, especially in terms of log loss. 
This is, however, not the case for RelNNs. 
In the RelNN model in \fig{RLR-RelNN}(b), the probability of maleness depends on $w_5*sigmoid(w_0 + w_1n_\const{U})$. 
$w_1$ controls the steepness of the slope of the sigmoid, $w_0$ moves the slope around, and $w_5$ controls the importance of the slope in the final prediction. 
With this model, as $n_\const{U}$ increases, the probability of maleness may converge to any number in the $[0,1]$ interval. Furthermore, this model enables learning of the critical zones.
\end{motivation}

\begin{motivation}
Suppose the underlying truth for the models in \fig{RLR-RelNN} is that males correspond to users who like at least $p$ action movies such that each movie has less than $q$ likes.
A typical RLR/MLN model cannot learn such a model because first it needs to learn something about movies (i.e. having less than $q$ likes), combine it with being \emph{action} and count them. 
However, this can be done using a RelNN (the hidden layer should contain a relational linear unit with atom $\pred{H}(\lv{m})$ and WFs \WF{w_1}{\pred{Likes}(\lv{u},\lv{m})} and \WF{w_2}{\true}). 
Thus, hidden layers increase the modelling power by enabling the model to learn generic rules and categorize the objects accordingly, then treat objects in different categories differently.
\end{motivation}

\begin{motivation}
\citeauthor{Kazemi:2014} (\citeyear{Kazemi:2014}) show how different types of existing explicit aggregators can be represented using RLR. 
However, some of those cases (e.g., noisy-or and mode $=t$) require two RLLs and two RALs, i.e. RelNNs.
\end{motivation}

\subsection{Learning latent properties directly}
Objects may contain latent properties that cannot be specified using general rules,
but can be learned directly from data during training.
Such properties have proved effective in many tasks (e.g., recommendation systems \cite{koren2009matrix}).
These properties can be also viewed as soft clustering the objects into different categories.
We call the latent properties specifically learned for each object \emph{numeric latent properties} and the general latent properties learned through WFs in RLLs \emph{rule-based latent properties} to distinguish between the two. In the RelNN in \fig{KDD15-RelNN}, for instance, $\pred{S1}(\lv{i})$, $\pred{S4}(\lv{u})$ and $\pred{S7}(\lv{u})$ are the outputs of RALs and so are rule-based latent properties whereas $\pred{N2}(\lv{b})$ and $\pred{N4}(\lv{a})$ are numeric latent properties.

Consider the models in \fig{RLR-RelNN} and let $\pred{Latent}(\lv{m})$ be a numeric latent property of the movies whose value is to be learned during training. 
One may initialize $\pred{Latent}(\lv{m})$ with random values for each movie and add a WF \WF{w}{\pred{Likes}(\lv{u},\lv{m}) * \pred{Latent}(\lv{m})} to the first RLL. 
As mentioned before, during the back propagation phase, an RLL provides the derivatives with respect to each of the inputs. 
This means for each grounding $\pred{Latent}(\ind{M})$, we will have $\Dout(\pred{Latent}(\ind{M}))$.
Therefore, these numeric latent values can also be updated during learning using gradient descent. 

\section{From ConvNet Primitives to RelNNs}
\citeauthor{niepert2016learning} (\citeyear{niepert2016learning}) and \citeauthor{pham2017column} (\citeyear{pham2017column}) explain why their relational learning models for graphs can be viewed as instances of ConvNet. We explain why RelNNs can also be viewed as an instance of ConvNets. Compared to prior work, we go into more details and provide more intuition. Such a connection offers the benefit of understanding and implementing RelNNs using ConvNet primitives.

The cells in input matrices of ConvNets (e.g., image pixels) have spatial correlation and spatial redundancy: cells closer to each other are more dependent than cells farther away. 
For instance if $M$ represents an input channel of an image, the dependence between $M[i,j]$ and $M[i+1,j+1]$ may be much more than the dependence between $M[i, j]$ and $M[i, j + 20]$.
To capture this type of dependency, convolution filters are usually small squared matrices.
Convolution filters contain tied parameters so different regions of the input matrices are treated identically.

For relational data, the dependencies in the input matrices (the relationships) are different: the cells in the same row or column (i.e. relationships of the same object) have higher dependence than the cells in different rows and columns (i.e. relationships of different objects). 
For instance for a matrix $L$ representing which users like which movies, the dependence between $L[i,j]$ and $L[i+1, j+1]$ (different people and movies) may be far less than the dependence between $L[i, j]$ and $L[i, j + 20]$ (same person and different movies).
A priori, all rows and columns of the input matrices are exchangeable.
Therefore, to adapt ConvNets for relational data, we need vector-shaped filters that are invariant to row and column swapping and better capture the relational dependence and the exchangeability assumption.

One way to modify ConvNets to be applicable to relational data is as follows. 
We consider relationships as inputs to the network. 
We apply vector-shaped convolution filters on the rows and columns of the relationship matrices. 
For instance for gender prediction from movie ratings, a convolution filter may be a vector with size equal to the number of the movies. 
The values of these filters can be learned during training (like ConvNet filters), or can be fixed in which case they correspond to observed properties (e.g., a vector representing which movies are action movies).
Convolving each filter with each matrix produces a vector. These vectors go through an activation layer and are used as a filter in the next layers. 
Besides convolving a filter with a matrix,  one can join two matrices and produce a new matrix for the next layer. Joining two matrices $R(x, m)$ and $T(m, a)$ produces a new matrix $S(x, a)$ where for some $X \in x$ and $A \in a$, $S(X, A)=\sum_{M \in m} R(X, M)*T(M,A)$. 
The filters in the next layers can be applied to either input matrices, or the ones generated in previous layers.
By constructing an architecture and training the network, the learned filters identify low and high level features from the input matrices. 

While these operations are modifications of ConvNet operations, they all correspond to the operations in our RLR/MLN perspective.
Vector-shaped filters correspond to including numeric latent properties in WFs.
Fixed filters correspond to including observed atoms with a single logvar in WFs.
Joining matrices corresponds to using two binary atoms in a WF.

\begin{figure*}[t]
\begin{center}
\includegraphics[width=\textwidth]{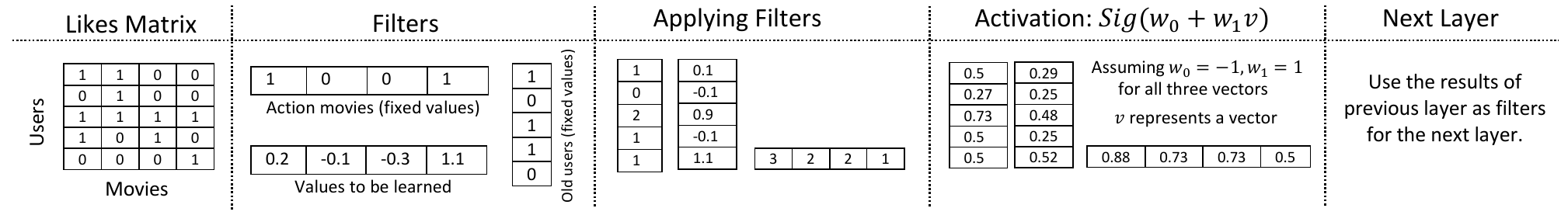}
\end{center}
\caption{An example of a layer of a RelNN demonstrated with ConvNet operations.}
\label{RelCNN}
\end{figure*}

\begin{example}
Suppose we want to predict the gender of users in a movie rating system based on the movies they like, whether the movie is action or not, and whether the user is old or not. \fig{RelCNN} shows one layer of a RelNN for this problem in terms of ConvNet operations. The binary relation $\pred{Likes}(\lv{u}, \lv{m})$ is considered as the input. Three filters have been applied to this matrix: one filter with fixed values corresponding to $\pred{Action}(\lv{m})$, one filter with fixed values corresponding to $\pred{Old(\lv{u})}$, and one filter to be learned corresponding to a numeric latent property $\pred{N}(\lv{m})$. Note that $\pred{Action}(\lv{m})$ and $\pred{N}(\lv{m})$ are row-shaped filters and $\pred{Old}(\lv{u})$ is column-shaped. Convolving these filters with $\pred{Likes}(\lv{u}, \lv{m})$ and sending the resulting vectors through an activation layer is equivalent to having an RLL with three relational linear units: the first unit contains:
\begin{center}
$\WF{\true}{w_0}$\\
$\WF{\pred{Likes}(\lv{u}, \lv{m})*\pred{Action}(\lv{m})}{w_1}$
\end{center}
and the other ones have $\pred{Old(\lv{u})}$ and $\pred{N}(\lv{m})$ instead of $\pred{Action}(\lv{m})$ respectively. The obtained vectors after activation can be used as filters for the next layers, or for making the final predictions.
\end{example}

\section{Empirical Results} \label{experiments-section-RelNNs}
In our experiments, we tend to answer these questions: \textbf{Q1:} how does RelNN's performance compare to other well-known relational learning algorithms, \textbf{Q2:} how numeric and rule-based latent properties affect the performance of the RelNNs, and \textbf{Q3:} how well RelNNs extrapolate to unseen cases and address the population size issue pointed out in \cite{Poole:2014} and discussed in Motivation~\ref{motiv1}. 

\textbf{Datasets:} We use three real-world datasets in our experiments. Our first dataset is the Movielens 1M dataset \cite{harper2015movielens} ignoring the actual ratings and only considering if a movie has been rated or not, and considering only action and drama genres. Our second dataset is from PAKDD15 gender prediction competition\footnote{\url{https://knowledgepit.fedcsis.org/contest/view.php?id=107}} but we only considered the $A$, $B$, and $C$ prefixes of the items and ignored the $D$ prefix because each $D$ prefix is on average seen by $\approx 1.5$ people. We also ignored the information within the sequence of the items viewed by each user. Our third dataset contains all Chinese and Mexican restaurants in Yelp dataset challenge\footnote{\url{https://www.yelp.com/dataset_challenge}} (ignoring the ones that have both Chinese and Mexican foods), and the task is to predict if a restaurant is Chinese or Mexican given the people who reviewed them, and whether they have fast food and/or seafoods or not.

\begin{table*}[t]
\scriptsize
  \caption{Performance of different learning algorithms based on accuracy, log loss, and MSE $\pm$ standard deviation on three different tasks. \emph{NA} means the method is not directly applicable to the prediction task/dataset. The best performing method is shown in bold. For models where standard deviation was zero, we did not report it in the table. 
  }
  \label{comparison-results}
  \centering
  \begin{tabular}{cccccccc}
    \toprule
    & & \multicolumn{6}{c}{Learning Algorithm}                   \\
    \cmidrule{3-8}
           Task        &                       Measure                                & Mean   & Matrix Factorization       & Collaborative Filtering & RDN-Boost    & RLR/MLN       & RelNN                 \\
    \midrule
	MovieLens 	  & \multicolumn{1}{c|}{Accuracy}     &  0.7088    & 0.7550 $\pm$ 0.0073  & 0.7510 &  0.7234  &   0.7073 $\pm$ 0.0067  &  \textbf{0.7902 $\pm$ 0.0051}\\
	Gender               & \multicolumn{1}{c|}{Log Loss}           &  0.8706     &  0.7318 $\pm$ 0.0073 & 0.7360  & 0.8143 &   0.8441 $\pm$ 0.0255   &  \textbf{0.6548 $\pm$ 0.0026} \\
                                  & \multicolumn{1}{c|}{MSE}              &  0.2065   & 0.1649 $\pm$ 0.0026 &  0.1675 & 0.1904  &   0.1987 $\pm$ 0.0067  &  \textbf{0.1459 $\pm$ 0.0009} \\
		    
       \midrule 		    
    
            PAKDD               & \multicolumn{1}{c|}{Accuracy}            &   0.7778   & NA   & 0.8806 &  0.7778    &   0.8145 $\pm$ 0.0388  & \textbf{0.8853 $\pm$ 0.0002}\\
          Gender            & \multicolumn{1}{c|}{Log Loss}                  &  0.7641   & NA  & 0.5135  &     0.8039   &    0.7224 $\pm$ 0.0548 & \textbf{0.5093 $\pm$ 0.0037}\\
                                  & \multicolumn{1}{c|}{MSE}                   &  0.1728     & NA &   0.1026 &  0.1842 &    0.1522 $\pm$ 0.0185 & \textbf{0.1009 $\pm$ 0.0008}\\
       \midrule 		    
       
             Yelp               & \multicolumn{1}{c|}{Accuracy}            &   0.6168   &   0.6154 $\pm$ 0.0041   & 0.6712 &  0.6156   &   0.6168 $\pm$ 0.000  & \textbf{0.6927 $\pm$ 0.0077}\\
          Business     & \multicolumn{1}{c|}{Log Loss}                  &  0.9604    &  0.9394 $\pm$ 0.0033 & 0.8757 & 0.9458  &    0.9435 $\pm$ 0.0034 & \textbf{0.8531 $\pm$ 0.0090}\\
           Prediction  & \multicolumn{1}{c|}{MSE}                   &  0.2364    &  0.2300  $\pm$ 0.0012  & 0.2084 & 0.2316    &    0.2309 $\pm$ 0.0010 & \textbf{0.2023 $\pm$ 0.0024}\\
       \midrule 		    
       
     MovieLens Age     & \multicolumn{1}{c|}{MSE}             &       156.0507   &  104.5967 $\pm$ 0.9978 &  90.8742 &  NA        &  156.0507 $\pm$ 0.000       & \textbf{62.7000 $\pm$ 0.7812}\\                
    \bottomrule
  \end{tabular}
\end{table*}

\begin{figure*}[t]
\begin{center}
\includegraphics[width=0.6\textwidth]{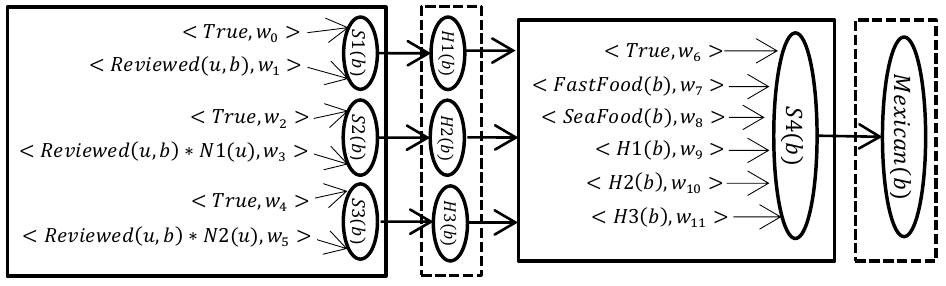}
\end{center}
\caption{RelNN structure used for Yelp dataset. $\pred{N}1(\lv{u})$ and $\pred{N}2(\lv{u})$ are numeric latent properties.}
\label{Yelp-RelNN}
\end{figure*}

\textbf{Learning algorithms and learning methodology:} Traditional machine learning approaches are not directly applicable to our problems. Also due to the large size of the datasets, some relational models do not scale to our problems. For instance, we tried Problog \cite{DeRaedt:2007}, but the software crashed after running for a few days even for simple models. The reason for scalability issues with many existing models (e.g., MLNs and Problog) is because they consider probabilistic (instead of neural) units and do inference and learning using EM. It took a day for an analogous EM-based implementation of RelNNs to learn a model for a synthetic dataset with 100 relations, whereas our current model (which is neural and uses back propagation) took under an hour for a million relations. We tried several baselines and only reported the ones that performed best on our datasets. \citeauthor{kazemi2017comparing} (\citeyear{kazemi2017comparing}) explain why each of the existing baselines for aggregation perform poorly and why this problem is still poorly understood. 

In our baselines, \emph{mean} corresponds to always predicting the mean of the training data. The \emph{matrix factorization} corresponds to the best performing matrix factorization model according to \citeauthor{kazemi2017comparing} (\citeyear{kazemi2017comparing})'s experiments: the relation matrix is factorized into object latent properties using the factorization technique in \cite{koren2009matrix}, then, using Weka  \cite{hall2009weka}, a linear/logistic regression model is learned over the latent and observed properties of the target objects. We also tried several other variants (including RESCAL's algorithm \cite{nickel2012factorizing}), but \citeauthor{kazemi2017comparing} (\citeyear{kazemi2017comparing})'s model was the best performing. The \emph{RDN-Boost} \cite{natarajan2012gradient} model can be viewed as an extension of random forests with ways to aggregate over multiple observations, thus enabling random forests to be applicable to relational data. The \emph{k-nearest neighbors collaborative filtering} model finds $k$ objects with observed labels that are similar to the target object in terms of the relationships they participate in, creates a feature as the weighted mean of the labels of the similar objects, and feeds this label along with other features of the target object to a classification/regression model. For MovieLens and Yelp datasets, collaborative filtering produces one feature and for PAKDD dataset, it produces three features, one for each item prefix. We used cosine similarity function. The value of $k$ was set using cross validation. Similar to matrix factorization, linear and logistic regression of Weka were used for classification and regression. 

For all RelNN and RLR/MLN models in our experiments, we used fixed structures (i.e. fixed set of WFs and connections among them) and learned the parameters using back-propagation with multiple restarts. The structure of the RelNN model for Movielens dataset is that of \fig{RLR-RelNN}(b) extended to include all ages and occupations with $2$ numeric latent properties, $3$ RLLs and $3$ RALs. The structure of the RelNN model used for PAKDD15 and Yelp datasets is as in \fig{KDD15-RelNN} and \fig{Yelp-RelNN} respectively. 
We leave the problem of learning these structures automatically from data as future work.
The structure of the RLR/MLN models are similar to the RelNN models but without hidden layers and numeric latent properties. Back propagation for the RLR/MLN case corresponds to the discriminative parameter learning of \cite{Huynh:2008}. To avoid numerical inconsistencies when applying our models to PAKDD15 dataset, we assumed there exist a male and a female in the training set who have viewed all items. For all experiments, we split the data into 80/20 percent train/test.

We imposed a Laplacian prior on all our parameters (weights and numeric latent properties). 
For classification, we further regularized our model predictions towards the mean of the training set using a hyper-parameter $\lambda$ as:
$Prob = \lambda * mean + (1 - \lambda) * (ModelSignal)$.
This regularization alleviates the over-confidence of the model and avoids numerical inconsistencies arising when taking the logs of the predictions. Note that this regularization corresponds to adding an extra layer to the network. 
We reported the \emph{accuracy} indicating the percentage of correctly classified instances, the \emph{mean squared errors (MSE)}, and the \emph{log loss}.
We conducted each experiment 10 times and reported the mean and standard deviation.

\textbf{Experiments:} Our experiments include predicting the gender for the Movielens and PAKDD15 datasets, predicting the age for Movielens dataset, and predicting the type of food for the Yelp restaurants dataset. 
Even though the age is divided into 7 categories in the MovieLens dataset, we assume it is a continuous variable to see how RelNNs perform in predicting continuous variables. 
To make the categories more realistic, for people in age category/interval [i, j], we assume the age is $(i+j)/2$, i.e. the mean of the interval. 
For the first and last categories, we used $16$ and $60$ as the ages.

Table~\ref{comparison-results} compares RelNNs with other well-known relational learning algorithms as well as a baseline. 
It can be viewed from the table how RelNNs outperform well-known relational learning models in terms of all three performance metrics. Note that all baselines use the same observed features. Manual feature engineering from relations is very difficult, so the challenge for the models is to extract useful features from relations. As explained in \cite{kazemi2017comparing}, the problem with the baselines is with their lower modeling power and inappropriate implicit assumptions, and not with engineering features.
The results in Table~\ref{comparison-results} answer \textbf{Q1}.

\begin{figure}[t]
\begin{center}
\includegraphics[width=\columnwidth]{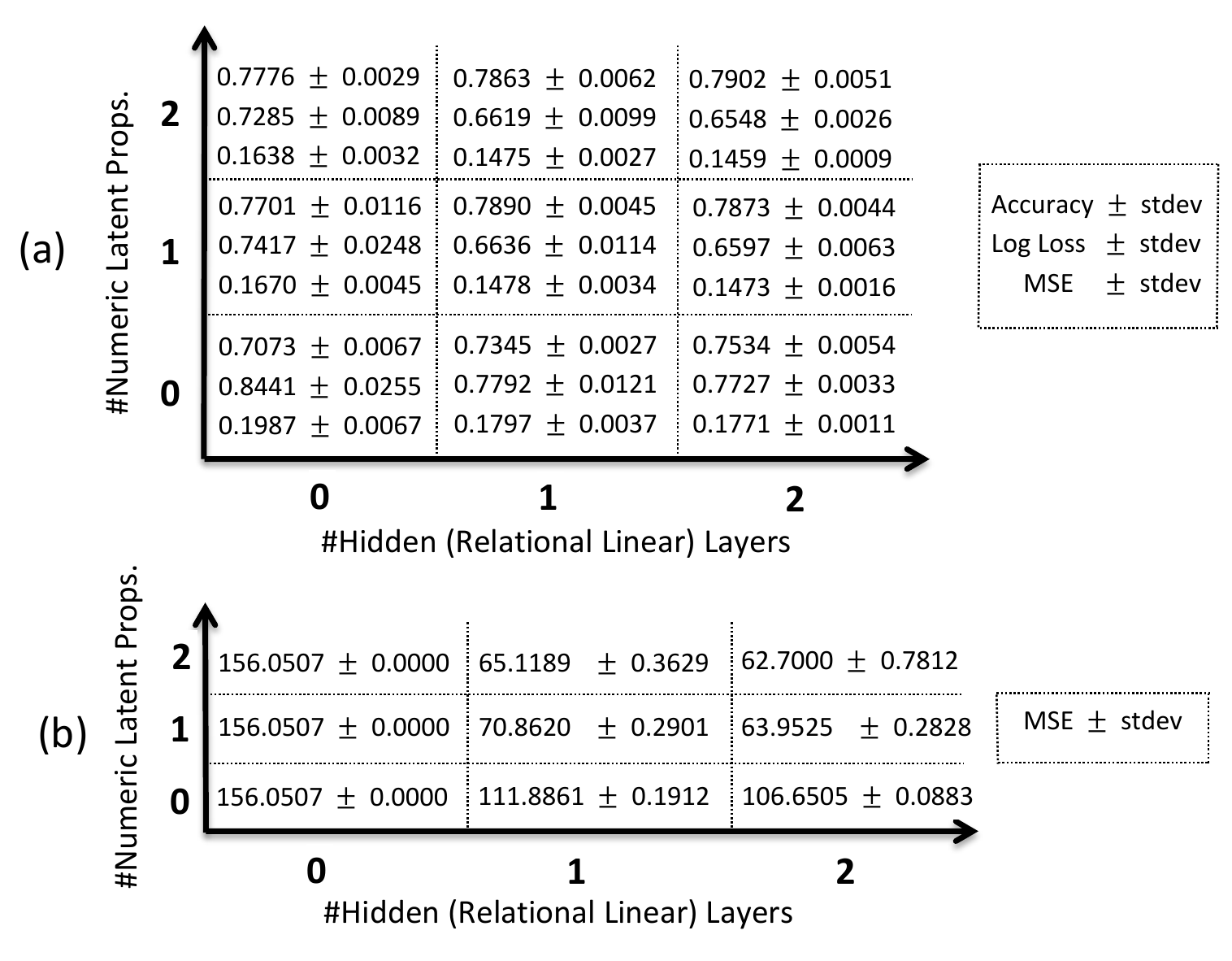}
\end{center}
\caption{Predicting (a) gender and (b) age on MovieLens using RelNNs with different number of numeric latent properties and hidden layers.}
\label{GenderResults}
\end{figure}

For \textbf{Q2}, we changed the number of hidden layers and numeric latent properties in RelNN to see how they affect the performance. 
The obtained results for predicting the gender and age in the Movielens dataset can be viewed in \fig{GenderResults}. 
The results show that both hidden layers and numeric latent properties (especially the first one of each) have a great impact on the performance of the model. 
When hidden layers are added, as we conjectured in motivation~\ref{motiv1}, the log loss and MSE improve substantially as the over-confidence of the model decreases. 
Note that adding layers only adds a constant number of parameters, but adding $k$ numeric latent properties adds $k * |\pop{m}|$ parameters. In \fig{GenderResults}, a RelNN with 2 hidden layers and 1 numeric latent property has many fewer parameters than a MLN/RLR with no hidden layers and 2 numeric latent properties, but outperforms it.

\begin{figure}[t]
\begin{center}
\includegraphics[width=0.8\columnwidth]{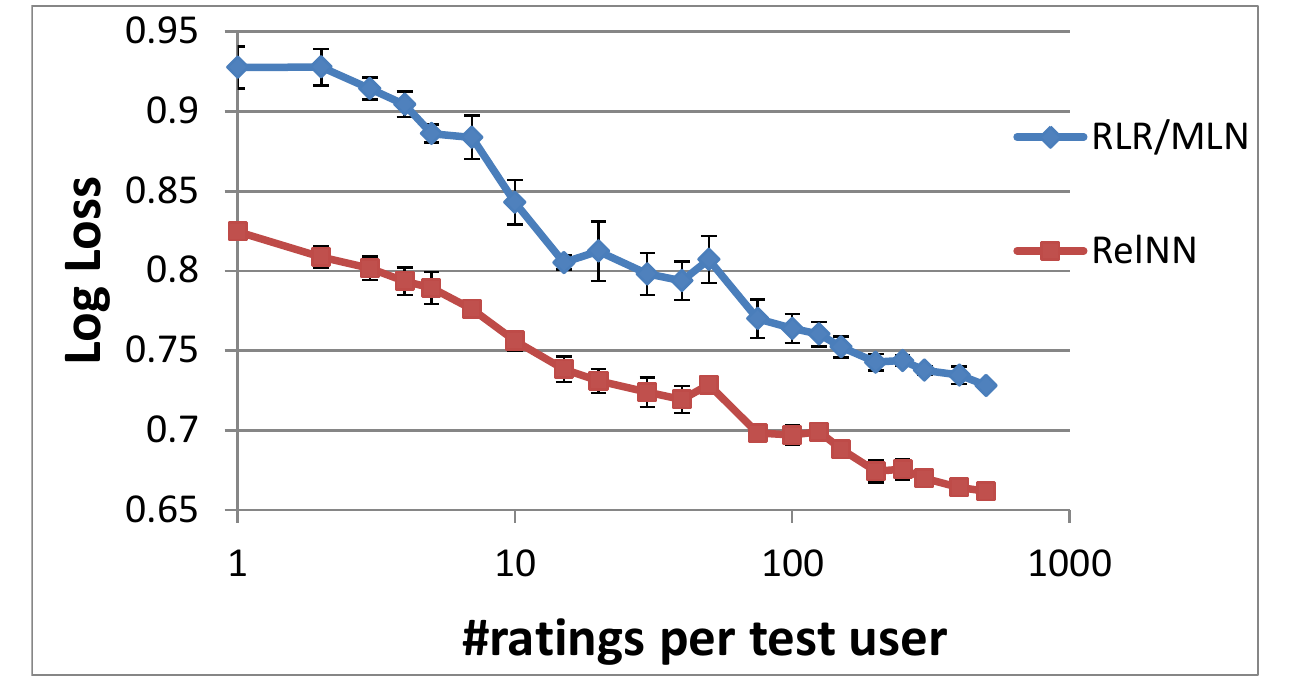}
\end{center}
\caption{Results on predicting the gender when we only use the first $k$ (on the x-axis) ratings of the test users and use all ratings in train set for learning.}
\label{Extrapolation1}
\end{figure}

For \textbf{Q3}, we conduct two experiments: 1- we train a RelNN on a large population, and test it on a small population, and 2- we train a RelNN on a small population and test it on a large population.
The first experiment can be also seen as how severely each model suffers from the cold start problem. 
The second experiment can be also seen as how well these models extrapolate to larger populations.
For the first experiment, we trained two models for predicting the gender of the users in the MovieLens dataset both containing one numeric latent property, but one containing no hidden layers and the other containing one hidden layer. 
We only gave the first $k$ ratings of the test users to the model and recorded the log loss of the two models. We did this for different values of $k$ and plotted the results. Obtained results can be viewed in \fig{Extrapolation1}.
Note that since in MovieLens dataset all users have rated at least 20 movies, the model has not seen a case where a user has rated less than $20$ movies. That is, the model has been trained on large population sizes ($\geq 20$). In this experiment, the cases in \fig{Extrapolation1} where $k < 20$ correspond to small populations (and cold start).

When $k=0$ (i.e. we ignored all ratings of the test users), both models had almost the same performance ($log loss \approx -0.9176$)\footnote{This is not shown in the diagram in \fig{Extrapolation1} as the x-axis of the diagram is in log scale.}. 
As soon as we add one rating for the test users, the performance of RelNN substantially improves, but the performance of the RLR/MLN model is not affected much.
According to the plot, the gap between the two models is more when the test users have fewer ratings (i.e. unseen cases), but it gets less and becomes steady as the number of ratings increases and becomes closer to the number of ratings in the train set. 

\begin{figure}[t]
\begin{center}
\includegraphics[width=0.8\columnwidth]{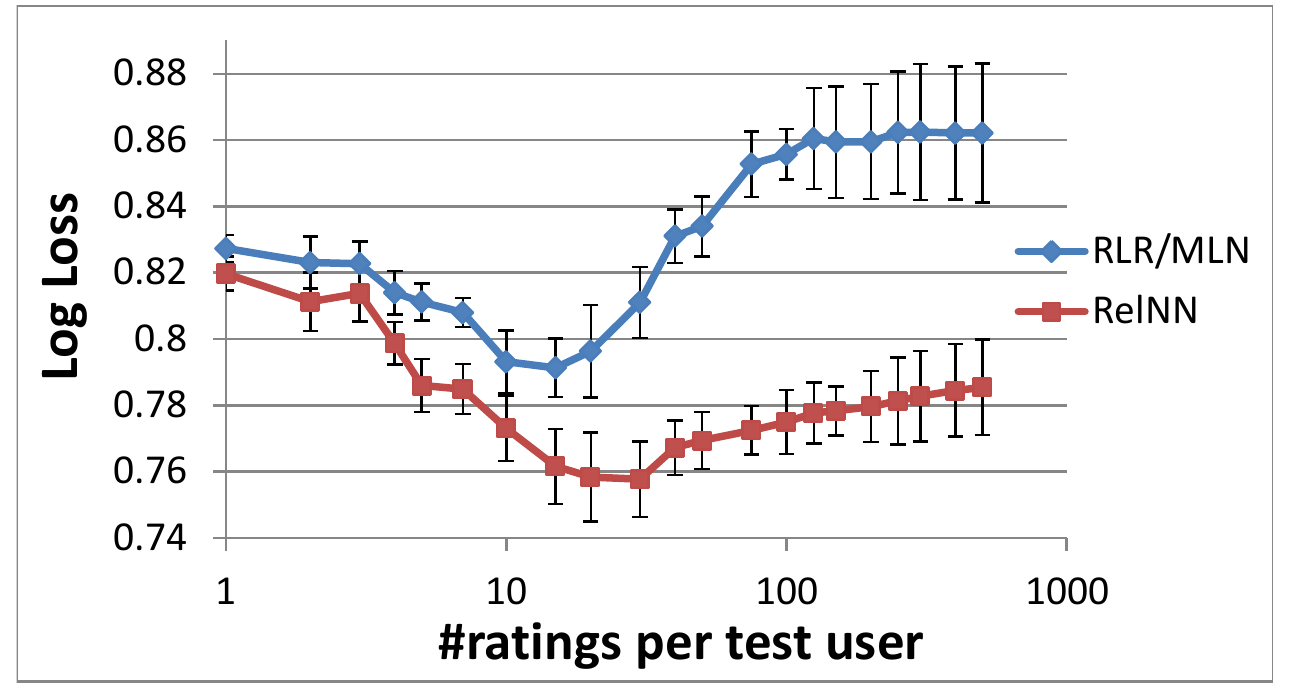}
\end{center}
\caption{Results on predicting the gender when we only use the first $k$ (on the x-axis) ratings of the test users and use $0\leq r \leq20$ ratings of each user ($r$ is generated randomly for each user) in the train set for learning.}
\label{Extrapolation2}
\end{figure}

For the second experiment, for each user in the train set we kept only the first $r$ ratings where $0\leq r \leq 20$ was generated randomly for each user. 
For the new dataset, we repeated the previous experiment and obtained the diagram in \fig{Extrapolation2}. 
It can be viewed in this diagram that the RLR/MLN model works best when we keep the first $15$ ratings of the test users (i.e. $k=15$), but for higher values of $k$, its performance starts to deteriorate. 
The performance of the RelNN model, on the other hand, improves even until $k=30$. 
After this point as $k$ increases, the performance of both models deteriorates, but the gap between the two models becomes much more for larger values of $k$, as the RLR/MLN model becomes over-confident.
These results validate the hypothesis in \cite{Poole:2014} that as the population size grows, RLR/MLN becomes over-confident and predicts with probabilities close to $0$ and $1$. The results also show how RelNNs address this issue and validate our Motivation~\ref{motiv1}.

\section{Related Work}
Recently, there has been a great body of research on learning from relational data using \emph{tensor factorization}. These methods learn embeddings for each object and each relationship. The probability of two objects participating in a relation is a simple function of the objects' and relation's embeddings (e.g., the sum of the element-wise product of the embeddings). Well-known tensor factorization works include \cite{nickel2012factorizing,bordes2013translating,socher2013reasoning,neelakantan2015compositional,trouillon2016complex} with a recent survey in \cite{nguyen2017overview}. Except RESCAL \cite{nickel2012factorizing}, all other works along this line only focus on predicting new relations from existing ones (cf. \cite{nickel2016review} section X.A). As described in \cite{kazemi2017comparing}, for aggregation problems studied in this work, RESCAL's proposal ends up memorizing the training labels and does not generalize to unseen cases. The same issue exists for other tensor factorization algorithms thus making them unsuitable for aggregation.

Besides tensor factorization models, several other relational learning models only focus on predicting relations and ignore properties, or, relying on embeddings, suffer from the same issues as tensor factorization algorithms. Examples of these works include path-constrained random walks (e.g., \cite{lao2010relational,lao2011random}), and several hybrid approaches (e.g., \cite{das2016chains,rocktaschel2017end,yang2017differentiable,wang2017relational}).

Several works on predicting object properties based on deep learning consider only a subset of relational data such as set data \cite{zaheer2017deep} or graph data \cite{pham2017column} (where there can be multiple relationships among objects, but all objects belong to the same class). Several works on predicting object properties are based on recurrent neural networks (RNNs) (see e.g., \cite{uwents2005classifying,moore2017deep}). \citeauthor{zaheer2017deep} (\citeyear{zaheer2017deep}) show that RNN based methods do not perform well for set data (a restricted form of relational data) as RNN-based models consider the input data to be a sequence and do not generalize to unseen orders. 

When a target variable depends on a population, the \emph{rule combining} \cite{kersting2001adaptive,natarajan2010exploiting} approaches learn a distribution $D$ for the target given only one object in the population, and then combine these distributions using an explicit aggregator (e.g., \emph{mean} or \emph{noisy-OR}). These models have three limitations compared to RelNNs: 1- they correspond to only two (non-flexible) layers, one for learning $D$ and one for combining the predictions, 2- since $D$ is learned separately for each object in the population, the interdependence among these objects is ignored, and 3- they rely on explicit aggregation function rather than learning the aggregator from the data. 

\emph{Predicate invention} approaches \cite{kemp2006learning,kok2007statistical} cluster the objects, their properties, and relationships such that the values of the target(s) depends on their clustering.
Such a clustering can be implemented as one layer of a RelNN. Since latent variables in these works are probabilistic (rather than neural as in RelNNs), these works typically result in expensive models such that in practice clustering is limited to hard rather than soft clusters, and they have been applied to small domains with around $10k$ observations (which is far less than, say, the $1M$ observations in our datasets).

There are also several other works for learning deep networks from relational data, but they are all limited in terms of modelling power or scalability. The works in \cite{garcez2012neural,francca2014fast,serafini2016logic}, for instance, propositionalize the data and miss the chance to learn specifically about objects. We show in our experiments that learning about objects substantially improves performance. \citeauthor{lodhi2013deep} (\citeyear{lodhi2013deep}) first learns features from relational data then feeds it into a standard neural network, thus learning features and the model independently. The same issue exists with methods such as DeepWalk \cite{perozzi2014deepwalk} which first learn embeddings for nodes regardless of the prediction task, and then use the embeddings for making predictions. \citeauthor{vsourek2015lifted} (\citeyear{vsourek2015lifted})'s models are the closest proposals to RelNNs, but RelNNs are more flexible in terms of adding new types of layers in a modular way.
These works are also limited in one or more of the following ways: 1- the model is limited to only one input relation, 2- the structure of the model is highly dependent on the input data, 3- the model allows for only one hidden layer, 4- the model cannot learn hidden object properties through general rules, or 5- the model does not scale to large domains.

\section{Conclusion}
In this paper, we developed a deep neural model for learning from relational data. We showed that our model outperforms several existing models on three relational learning benchmarks. Our work can be advanced in several ways.
The current implementation of our model is a fresh non-parallelized code. It could be speeded up by parallelization (e.g., by using TensorFlow \cite{TensorFlow} as backbone), by compilation relational operations to lower-level languages (similar to \cite{LRC2CPP}), or by using advanced database query operations. 
Learning the structure of the RelNN and the WFs automatically from data (e.g., by extending the structure learning algorithm in \cite{fatemi2016learning}), developing and adding relational versions of regularization techniques such as (relational) dropout \cite{dropout,kazemi2017comparing} and batch normalization \cite{ioffe2015batch} are other directions for future research.

\bibliography{MyBib}
\bibliographystyle{AAAI}

\end{document}